\documentclass[letterpaper]{article} 
\usepackage{aaai2026}  
\usepackage{times}  
\usepackage{helvet}  
\usepackage{courier}  
\usepackage[hyphens]{url} 
\usepackage{graphicx} 
\usepackage{natbib} 
\usepackage{caption} 
\usepackage{algorithm}
\usepackage{algorithmic}
\usepackage{amsmath}
\usepackage{mathtools}
\usepackage{amssymb}
\usepackage{dsfont}
\usepackage{booktabs} 
\usepackage{subfig}
\usepackage{xcolor}
\usepackage{colortbl}
\usepackage{booktabs}
\usepackage{multirow}
\usepackage{adjustbox}
\usepackage{arydshln}
\usepackage{tikz}
\usetikzlibrary{tikzmark}

\usepackage{xcolor}
\usepackage{colortbl}
\newcommand\Video[1]{\colorbox[RGB]{224,235,247}{#1}}

\newcommand\InputName[1]{\colorbox[RGB]{231,220,250}{#1}}
\newcommand\neurips[1]{\colorbox[RGB]{226,225,238}{#1}}
\newcommand\acl[1]{\colorbox[RGB]{231,218,210}{#1}}
\newcommand\InitialReasoning[1]{\colorbox[RGB]{227,238,225}{#1}}

\definecolor{red}{RGB}{184,49,55}

\usepackage{newfloat}
\usepackage{listings}
\DeclareCaptionStyle{ruled}{labelfont=normalfont,labelsep=colon,strut=off}
\floatstyle{ruled}
\newfloat{listing}{tb}{lst}{}
\floatname{listing}{Listing}

\title{AAAI Press Formatting Instructions \\for Authors Using \LaTeX{} --- A Guide}
\title{Let's Think with Images Efficiently! An Interleaved-Modal Chain-of-Thought Reasoning Framework with Dynamic and Precise Visual Thoughts}
\author{
    Xu Liu\textsuperscript{\rm 1,\rm 2,\rm 3$\ast$}, Yongheng Zhang\textsuperscript{\rm 2}\thanks{~Equal Contribution.}, Qiguang Chen\textsuperscript{\rm 2}, Yao Li\textsuperscript{\rm 4}, Sheng Wang\textsuperscript{\rm 4}, Libo Qin\textsuperscript{\rm 1,\rm 2,\rm 3}
}
\affiliations{
   \textsuperscript{\rm 1}Institute of Computing and Intelligence, Harbin Institute of Technology, Shenzhen \\
    \textsuperscript{\rm 2}School of Computer Science and Engineering, Central South University\\
    \textsuperscript{\rm 3}Text Computing and Cognitive Intelligence Ministry of Education Engineering Research Center, Guizhou University\\
   \textsuperscript{\rm 4}Shanghai Aviation Electric Co., Ltd, Aviation Industry Corporation of China, Shanghai
}

\usepackage{bibentry}

\begin{document}

\maketitle

\begin{abstract}
Recently, Interleaved-modal Chain-of-Thought (ICoT) reasoning has achieved remarkable success by leveraging both multimodal inputs and outputs, attracting increasing attention. While achieving promising performance, current ICoT methods still suffer from two major limitations: (1) \textit{Static Visual Thought Positioning}, which statically inserts visual information at fixed steps, resulting in inefficient and inflexible reasoning; and (2)~\textit{Broken Visual Thought Representation}, which involves discontinuous and semantically incoherent visual tokens.
To address these limitations, we introduce Interleaved-modal Chain-of-Thought reasoning with \textbf{D}ynamic \textbf{a}nd \textbf{P}recise Visual Thoughts (\textsc{DaP-ICoT}), which incorporates two key components:
(1) \textit{Dynamic Visual Thought Integration} adaptively introduces visual inputs based on reasoning needs, reducing redundancy and improving efficiency.
(2) \textit{Precise Visual Thought Guidance} ensures visual semantically coherent and contextually aligned representations.
Experiments across multiple benchmarks and models demonstrate that \textsc{DaP-ICoT} achieves state-of-the-art performance. In addition, \textsc{DaP-ICoT} significantly reduces the number of inserted images, leading to a 72.6\% decrease in token consumption, enabling more efficient ICoT reasoning.
\end{abstract}

\begin{links}
    \link{Code}{https://github.com/67L1/DaP-ICoT}
\end{links}

\section{Introduction}

In recent years, Multimodal Large Language Models (MLLMs)~\cite{achiam2023gpt,team2024gemini,qin2024large}, and Multimodal Chain-of-Thought (MCoT)~\cite{zhang2023multimodal,chen-etal-2024-m3cot}, have significantly advanced the reasoning capabilities of MLLMs across the complex real-world tasks~\cite{wei2025ad,chen2025towards}.
However, current MCoT mainly follows the traditional paradigm: cross-modal input, but reasoning output in the text modality, which limits the human’s ability to exploit modality complementarity, reducing its reasoning performance~\cite{fei2024video,menon2024whiteboard,wu2025vic}.
To address this limitation, researchers propose Interleaved-Modal Chain-of-Thought (ICoT) reasoning~\cite{hu2024visual,cheng2025comt,gao2025interleaved}. ICoT allows the model to receive multimodal input and simultaneously perform multimodal reasoning output, enabling visual thoughts to effectively convey the image information during reasoning~\citep{meng2023chain,shao2024visual, zhang2024wrong,li2025imagine,wang2025multimodal,chen2025ai4research}.
\begin{figure}[!t]
\centering
\includegraphics[width=\columnwidth]{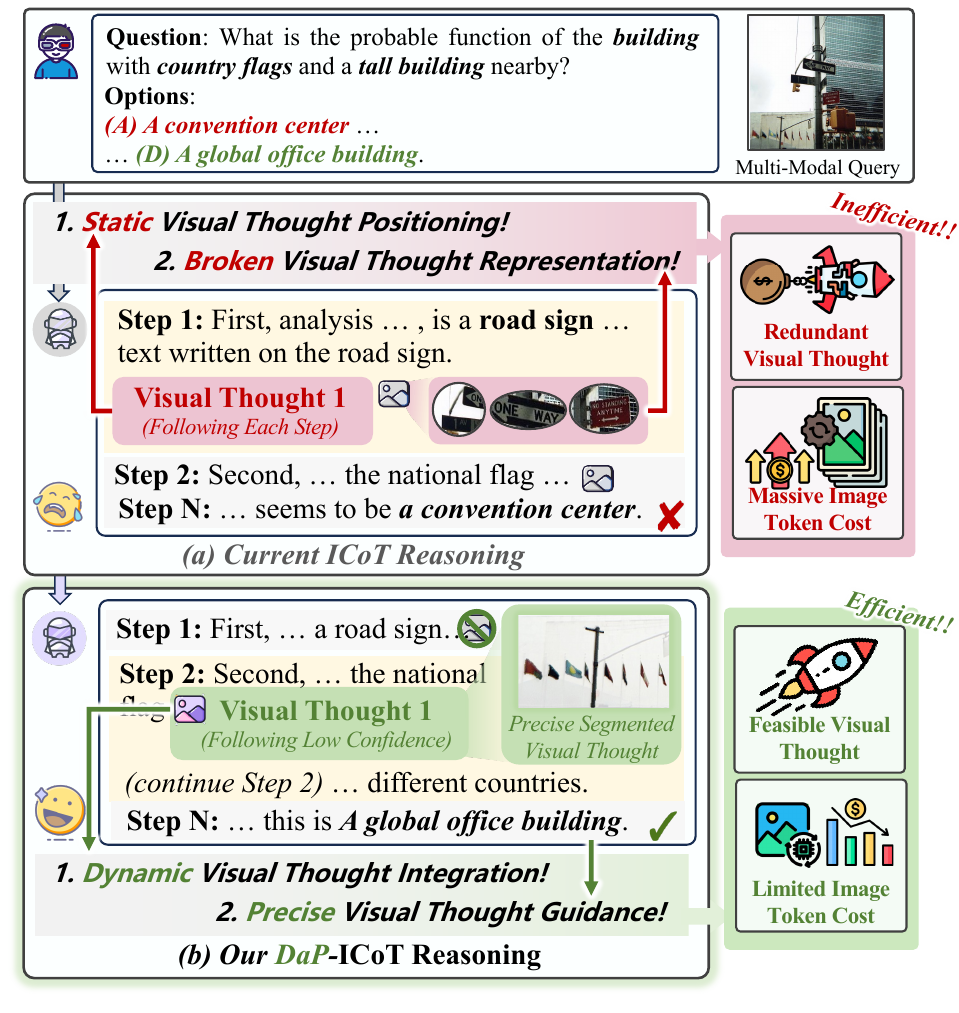}
\vspace{-4mm}
\caption{(a) Current ICoT: While supporting multimodal inputs and outputs, it suffers from \textit{Static Visual Thought Integration}, which requires the insertion of visual information after each step, and \textit{Broken Visual Thought Representation}, in which the inserted visual tokens are lack coherence, resulting in inefficient reasoning. (b) Our \textsc{DaP-ICoT}: It provides \textbf{\textit{Dynamic Visual Thought Integration}} and \textbf{\textit{Precise Visual Thought Guidance}}, enabling efficient reasoning.}
\vspace{-6mm}
\label{intro}
\end{figure}

Specifically, a growing body of research has focused on advancing ICoT reasoning to better convey visual thoughts.
\citet{cheng2025comt} employ segmentation tools to extract key areas and generate images that form visual thoughts, thereby significantly enhancing the capabilities of MLLMs for more advanced, human-like multimodal reasoning and visual operations across diverse tasks. \citet{hu2024visual} propose Sketchpad, a novel and effective sketching-based framework that enhances MLLM reasoning by enabling visual thought expression through intuitive figure drawing on a dynamic visual sketchpad interface. \citet{zhou2024image} propose Image-of-Thought visual prompting, a structured method for step-by-step visual rationale extraction that further enhances multimodal reasoning by effectively conveying internal visual thoughts in MLLMs through designed visual cues.
\citet{zhang2025vitcot} propose Video-Text Interleaved CoT reasoning, a cognitively aligned temporal video reasoning paradigm that interleaves visual and textual information to enhance video understanding and reasoning in MLLMs.
\citet{gao2025interleaved} further employ an attention-driven selection module to statically select some image tokens that convey visual thoughts in each step.

While achieving promising performance, as shown in Figure~\ref{intro}~(a), current ICoT methods for visual thought conveyance face two key challenges that impede their potential:
\begin{itemize}
    \item [(1)] \textbf{\textit{Static Visual Thought Positioning}}: Existing ICoT reasoning approaches insert visual thoughts statically after each textual rationale generation step, resulting in rigid thinking patterns, redunant visual information, and a significant increase in computational overhead.
    \item [(2)] \textbf{\textit{Broken Visual Thought Representation}}: Existing ICoT methods select discontinuous image tokens as visual thoughts, which are broken and lack coherence. Such broken visual thought impairs understanding and increases the risk of overlooking critical information.
\end{itemize}

Motivated by these challenges, we introduce an Interleaved-modal Chain-of-Thought reasoning framework with \textbf{D}ynamic \textbf{a}nd \textbf{P}recise Visual Thoughts (\textsc{DaP-ICoT}).  Specifically, to address the first challenge, as illustrated in Figure~\ref{intro}~(b), we propose \textbf{\textit{Dynamic Visual Thought Integration}}, which dynamically and adaptively leverages visual information in response to evolving reasoning needs. In contrast to statically processing all available visual data, \textsc{DaP-ICoT} selectively invokes visual modalities based on contextual demands, ensuring that the integration of visual cues is both timely and relevant. This dynamic mechanism enhances multimodal reasoning by reducing redundant computation and focusing on salient visual cues.
To address the second challenge, we introduce \textbf{\textit{Precise Visual Thought Guidance}}, which emphasizes the integration of semantically coherent and contextually relevant visual information. Instead of relying on broken cues, it employs precise visual representations that encapsulate complete semantics, ensuring tight alignment between visual input and the reasoning trajectory. This precision-oriented design safeguards conceptual consistency and enhances interpretability and reasoning accuracy.
Such two modules together minimize unnecessary visual information and better capture key visual thoughts, enabling more efficient ICoT reasoning.

Experiments conducted on multiple widely used benchmarks and MLLMs demonstrate that \textsc{DaP-ICoT} consistently outperforms baselines. Further in-depth analysis reveals that \textsc{DaP-ICoT} significantly reduces the number of image insertions, resulting in a 72.6\% reduction in token consumption compared to the current ICoT approach. These results highlight its effective reasoning, balancing efficiency and effectiveness in multimodal understanding.

Our contributions can be summarized as follows:

\begin{itemize}
    \item [(1)] We highlight two drawbacks in the existing ICoT paradigm: \textit{Static Visual Thought Positioning} that induces inefficient reasoning, and \textit{Broken Visual Thought Representation} that undermines coherent visual thoughts.

    \item [(2)] We introduce the Interleaved-modal Chain-of-Thought reasoning with \textbf{D}ynamic \textbf{a}nd \textbf{P}recise Visual Thoughts (\textsc{DaP-ICoT}) to address drawbacks in ICoT, which incorporates modules: \textit{Dynamic Visual Thought Integration} and \textit{Precise Visual Thought Guidance}.

    \item [(3)]  Extensive experiments demonstrate that \textsc{DaP-ICoT} achieves the state-of-the-art performance. In addition, \textsc{DaP-ICoT} effectively reduces the number of inserted images and achieves a 72.6\% reduction in token consumption, enabling more efficient reasoning.
    
\end{itemize}

\section{\textsc{DaP-ICoT} Reasoning}
\begin{figure*}[t]
\centering
\includegraphics[width=0.96\textwidth]{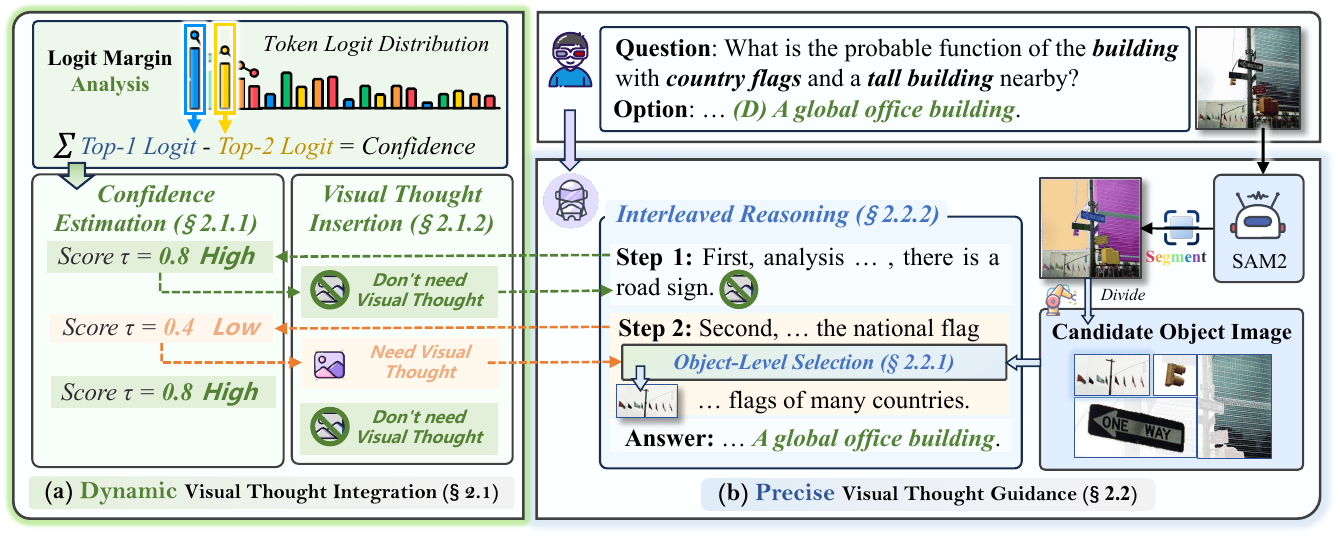} 
\caption{An overview of Interleaved-modal Chain-of-Thought reasoning with \textbf{D}ynamic \textbf{a}nd \textbf{P}recise Visual Thoughts~(\textsc{DaP-ICoT}), including \textit{Dynamic Visual Thought Integration} ($\S \ref{DVTI}$), and \textit{Precise Visual Thought Guidance}~($\S \ref{PVTG}$).}
\label{fig2}
\end{figure*}

 In this work, we introduce the Interleaved-modal Chain-of-Thought reasoning with \textbf{D}ynamic \textbf{a}nd \textbf{P}recise Visual Thoughts (\textsc{DaP-ICoT}) to address the inefficiencies in previous ICoT approaches.
 Specifically, as shown in Figure~\ref{fig2}, \textsc{DaP-ICoT} comprises \textbf{\textit{Dynamic Visual Thought Integration}}~($\S \ref{DVTI}$) and \textbf{\textit{Precise Visual Thought Guidance}}~($\S \ref{PVTG}$).

\subsection{Dynamic Visual Thought Integration}
\label{DVTI}
To fill the gap of the static visual thought positioning in existing ICoT reasoning methods, we introduce \textbf{\textit{Dynamic Visual Thought Integration}} (DVTI), as shown in Figure~\ref{fig2} (a), a confidence-aware strategy that dynamically decides whether visual thought should be integrated during reasoning, based on the confidence of MLLMs.

\vspace{3mm}
\noindent
\textbf{2.1.1 Confidence Estimation via Logit Margin Analysis}
\vspace{1mm}

Formally, given a generated textual rationale $T_t$ at reasoning step $t$, we estimate the model’s confidence by analyzing token-level logit differentials during generation. This allows us to internally assess certainty without requiring external calibration.
At each decoding position $i$, we define the local confidence $\delta_i$ as the difference between the highest and the second-highest predicted logits:
\begin{equation}
\delta_i = \ell_{i, w^{(1)}} - \ell_{i, w^{(2)}}, \quad \forall i \in {1, \ldots, |T_t|},
\end{equation}
where $\ell_{i, w^{(1)}}$ and $\ell_{i, w^{(2)}}$ denote the top-1 and top-2 logits predicted at position $i$, respectively. This margin reflects the model's decisiveness at each token position, larger values indicate stronger confidence in the token selection.

To obtain a more robust and reliable overall confidence score for the entire rationale $T_t$, we aggregate the local margins by computing their mean value across positions:
\begin{equation}
C_t = \frac{1}{|T_t|} \sum_{i=1}^{|T_t|} \delta_i = \frac{1}{|T_t|} \sum_{i=1}^{|T_t|} \left( \ell_{i, w^{(1)}} - \ell_{i, w^{(2)}} \right),
\end{equation}
where $C_t$ represents the average confidence across decoding steps, with each $\delta_i$ quantifying the model’s certainty at position $i$ based on the logit gap between its top two predicted tokens, reflecting how decisively the selection of MLLMs.

\noindent
\textbf{2.1.2 Visual Thought Insertion Guided by Confidence}
\vspace{1mm}

Based on the computed confidence score $C_t$, we apply a thresholding mechanism to determine whether visual input is necessary for the next step. The determinant is as follows:
\begin{equation}
I_{t+1} =
\begin{cases}
I^{\text{vision}}, & \text{if } C_t < \tau \\
\emptyset, & \text{otherwise}
\end{cases}
\end{equation}
where $ I_{t+1} $ denotes the visual thought input used at the $(t+1)$-th reasoning step. If the confidence $C_t$ falls below the predefined threshold $\tau$, the model is prompted to incorporate visual context, retrieved via the \textit{Precise Visual Thought Guidance} (§\ref{PVTG}). Otherwise, no visual input is provided ($\emptyset$), allowing for proceeding based on textual reasoning.

DVTI is capable of reducing redundant information through dynamic visual information integration, thereby decreasing token processing and improving efficiency.

\subsection{Precise Visual Thought Guidance}
\label{PVTG}
To fill the gap of broken and semantically incoherent visual inputs in existing ICoT methods, we propose \textbf{\textit{Precise Visual Thought Guidance}} (PVTG), as shown in Figure~\ref{fig2}~(b), which facilitates fine-grained object-level visual selection through cross-modal semantic relevance matching.

\vspace{2mm}
\noindent
\textbf{2.2.1 Object-Level Selection via Cross-Modal Relevance}
\vspace{1mm}

In the first, we apply the Segment Anything Model 2 (SAM2)~\cite{sam2} to the original image $I_{\text{ori}}$ for object-level segmentation. This identifies multiple distinct object regions, each corresponding to semantically meaningful visual content. We then extract sub-images of objects, forming a pool of candidate object-centric visual inputs. Unlike patch-level image tokens, these object-centric sub-images preserve semantic information. Specifically:
\begin{equation}
\mathcal{O} = {O_1, O_2, \ldots, O_N},
\end{equation}
where each $O_i$ denotes a complete and semantically coherent object sub-image extracted from the original image $I_{\text{ori}}$.

When \textit{Dynamic Visual Thought Integration} (§\ref{DVTI}) identifies the need for visual input at step $t$, we compute the semantic relevance between the textual rationale $T_t$ and each candidate object image $O_i \in \mathcal{O}$ via cross-modal attention using a similarity function $f_{\text{attn}}(\cdot, \cdot)$:
\begin{equation}
s_i = f_{\text{attn}}(T_t, O_i),
\end{equation}
where $s_i$ denotes the attention-based similarity between rationale $T_t$ and object image $O_i$ \cite{gao2025interleaved}.
We then select the most relevant object image with the highest score:
\begin{equation}
\hat{O} = \underset{O_i \in \mathcal{O}}{\operatorname{argmax}} \ s_i.
\end{equation}
where $\hat{O}$ denotes the object image that exhibits the strongest semantic alignment with the current textual rationale. 

\vspace{2mm}
\noindent
\textbf{2.2.2 Interleaved Reasoning with Aligned Visual Inputs}
\vspace{1mm}

Instead of treating $\hat{O}$ as a standalone input, we interleave it into the reasoning process by embedding it within the textual rationale $\mathcal{R}_t$, forming a multimodal reasoning sequence:
\begin{equation}
\mathcal{R}_{t \rightarrow v} = \mathcal{R}_t \oplus \hat{O},
\end{equation}
where $\oplus$ denotes the operation of embedding the selected object image $\hat{O}$ into the textual reasoning sequence $\mathcal{R}_t$, specifically by inserting the image token after the previously generated textual rationale.

The model then continues the next reasoning step based on this interleaved multimodal input:
\begin{equation}
\mathcal{R}_{t+1} = \underset{\mathcal{R}}{\operatorname{argmax}} \ P(\mathcal{R} \ |Q, \ \mathcal{R}_{t \rightarrow v}, P_{t \rightarrow v}),
\end{equation}
where $ Q $ denotes the question, and $P_{t \rightarrow v}$ represents the prompt constructed for interleaved reasoning.

Through targeted selection and structured interleaving, PVTG provides precise visual information that preserves semantic coherence, reduces noise, and enhances the effectiveness and interpretability of the multimodal reasoning.

\begin{table*}[t]
\setlength{\tabcolsep}{16pt}
\centering
\begin{adjustbox}{width=0.98\textwidth}
\begin{tabular}{ll cccccc}
\toprule

\multirow{3}{*}{\textbf{Models}} & \multirow{3}{*}{\textbf{Methods}} & \multicolumn{2}{c}{\textbf{M$^3$CoT}} & \multicolumn{2}{c}{\textbf{ScienceQA}} & \multicolumn{2}{c}{\textbf{MME}} \\

\cmidrule{3-8}

& & \textbf{\shortstack{0-Shot \\ Acc. $\uparrow$}} & \textbf{\shortstack{1-Shot \\ Acc. $\uparrow$}} & \textbf{\shortstack{0-Shot \\ Acc. $\uparrow$}} & \textbf{\shortstack{1-Shot \\ Acc. $\uparrow$}} & \textbf{\shortstack{0-Shot \\ Score $\uparrow$}} & \textbf{\shortstack{1-Shot \\ Score $\uparrow$}} \\

\rowcolor[rgb]{ .965, .985, .995 }

\midrule

\rowcolor[rgb]{ .965, .985, .995 }
& \textsc{Direct} & 22.5 & 23.8 & 43.1 & 43.4 & 724.2 & 942.9 \\
\rowcolor[rgb]{ .965, .985, .995 }
& \textsc{MMCoT} \Video{TMLR 2024} & 26.0 & 28.0 & 46.2 & 48.9 & 435.8 & 661.2 \\
\rowcolor[rgb]{ .965, .985, .995 }
& \textsc{DDCoT} \neurips{NeurIPS 2023} & 29.8 & 30.3 & 47.4 & 48.4 & 725.9 & 953.8 \\
\rowcolor[rgb]{ .965, .985, .995 }
& \textsc{SCAFFOLD} \acl{ACL 2025} & 31.0 & 31.2 & 48.6 & 50.7 & 388.1 & 634.5 \\
\rowcolor[rgb]{ .965, .985, .995 }
& \textsc{CCoT} \InputName{CVPR 2024} & 25.1 & 26.3 & 42.8 & 44.3 & 366.1 & 487.9 \\
\rowcolor[rgb]{ .965, .985, .995 }
& \textsc{ICoT} \InputName{CVPR 2025} & 26.1 & 32.1 & 44.5 & 45.3 & 794.8 & 928.9 \\
\rowcolor[rgb]{ .965, .985, .995 }
\multirow{-7.5}{*}{\shortstack{\textbf{\textit{Chameleon-7B}} \\ \cite{team2024chameleon}}} & \InitialReasoning{\textbf{\textsc{DaP-ICoT}}} & \textbf{41.0} & \textbf{41.9} & \textbf{57.1} & \textbf{62.9} & \textbf{832.3} & \textbf{1013.0} \\

\midrule

\rowcolor[rgb]{ .970, .978, .999 }
& \textsc{Direct} & 23.2 & 25.7 & 21.9 & 23.5 & 975.3 & 1079.8 \\
\rowcolor[rgb]{ .970, .978, .999 }
& \textsc{MMCoT} \Video{TMLR 2024} & 29.6 & 30.4 & 42.4 & 43.1 & 1140.2 & 1277.4 \\
\rowcolor[rgb]{ .970, .978, .999 }
& \textsc{DDCoT} \neurips{NeurIPS 2023} & 25.4 & 26.5 & 29.3 & 31.0 & 736.7 & 991.8 \\
\rowcolor[rgb]{ .970, .978, .999 }
& \textsc{SCAFFOLD} \acl{ACL 2025} & 26.6 & 28.3 & 37.7 & 39.8 & 1170.3 & 1264.3 \\
\rowcolor[rgb]{ .970, .978, .999 }
& \textsc{CCoT} \InputName{CVPR 2024} & 34.2 & 35.5 & 33.8 & 35.7 & 1294.3 & 1349.5 \\
\rowcolor[rgb]{ .970, .978, .999 }
& \textsc{ICoT} \InputName{CVPR 2025} & 34.6 & 35.0 & 41.7 & 46.7 & 1331.6 & 1421.6 \\
\rowcolor[rgb]{ .970, .978, .999 }
\multirow{-7.5}{*}{\shortstack{\textbf{\textit{LLaVA-V1.5-7B}} \\ \cite{liu2023visual}}} & \InitialReasoning{\textbf{\textsc{DaP-ICoT}}} & \textbf{36.3} & \textbf{37.6} & \textbf{50.4} & \textbf{51.1} & \textbf{1386.7} & \textbf{1526.9} \\

\midrule

\rowcolor[rgb]{ .970, .978, .999 }
& \textsc{Direct}  & 24.6 & 25.5 & 29.7 & 34.0 & 995.4 & 1118.5 \\
\rowcolor[rgb]{ .970, .978, .999 }
& \textsc{MMCoT} \Video{TMLR 2024} & 32.1 & 32.9 & 56.1 & 58.3 & 1078.1 & 1224.3 \\
\rowcolor[rgb]{ .970, .978, .999 }
& \textsc{DDCoT} \neurips{NeurIPS 2023} & 32.9 & 33.8 & 39.3 & 41.8 & 800.9 & 1034.1 \\
\rowcolor[rgb]{ .970, .978, .999 }
& \textsc{SCAFFOLD} \acl{ACL 2025} & 31.9 & 33.2 & 41.7 & 43.8 & 1231.2 & 1389.9 \\
\rowcolor[rgb]{ .970, .978, .999 }
& \textsc{CCoT} \InputName{CVPR 2024} & 30.1 & 32.0 & 45.0 & 46.3 & 1249.3 & 1442.3 \\
\rowcolor[rgb]{ .970, .978, .999 }
& \textsc{ICoT} \InputName{CVPR 2025} & 37.0 & 37.9 & 54.6 & 54.8 & 1405.4 & 1523.8 \\
\rowcolor[rgb]{ .970, .978, .999 }
\multirow{-7.5}{*}{\shortstack{\textbf{\textit{LLaVA-V1.5-13B}} \\ \cite{liu2023visual}}} & \InitialReasoning{\textbf{\textsc{DaP-ICoT}}} & \textbf{39.4} & \textbf{41.8} & \textbf{60.3} & \textbf{62.7} & \textbf{1556.3} & \textbf{1726.3} \\

\midrule

\rowcolor[rgb]{ .961, .985, .970 }
& \textsc{Direct}  & 14.4 & 24.5 & 64.3 & 65.2 & 641.5 & 741.3 \\
\rowcolor[rgb]{ .961, .985, .970 }
& \textsc{MMCoT} \Video{TMLR 2024} & 14.9 & 22.4 & 65.6 & 67.3 & 1102.8 & 1304.7 \\
\rowcolor[rgb]{ .961, .985, .970 }
& \textsc{DDCoT} \neurips{NeurIPS 2023} & 37.9 & 39.3 & 65.8 & 68.4 & 800.6 & 967.0 \\
\rowcolor[rgb]{ .961, .985, .970 }
& \textsc{SCAFFOLD} \acl{ACL 2025} & 40.3 & 43.6 & 66.7 & 69.4 & 1344.8 & 1536.2 \\
\rowcolor[rgb]{ .961, .985, .970 }
& \textsc{CCoT} \InputName{CVPR 2024} & 20.2 & 37.7 & 64.2 & 66.5 & 761.6 & 867.5 \\
\rowcolor[rgb]{ .961, .985, .970 }
& \textsc{ICoT} \InputName{CVPR 2025} & 35.8 & 37.3 & 60.4 & 67.0 & 941.9 & 1453.9 \\
\rowcolor[rgb]{ .961, .985, .970 }
\multirow{-7.5}{*}{\shortstack{\textbf{\textit{Qwen2-VL-2B}} \\ \cite{wang2024qwen2}}} & \InitialReasoning{\textbf{\textsc{DaP-ICoT}}} & \textbf{47.3} & \textbf{51.0} & \textbf{68.4} & \textbf{73.6} & \textbf{1378.9} & \textbf{1862.4} \\

\midrule

\rowcolor[rgb]{ .961, .985, .970 }
& \textsc{Direct} & 33.0 & 35.0 & 70.9 & 71.2 & 1599.3 & 1641.3 \\
\rowcolor[rgb]{ .961, .985, .970 }
& \textsc{MMCoT} \Video{TMLR 2024} & 44.4 & 47.5 & 70.8 & 73.8 & 1602.5 & 1874.3 \\
\rowcolor[rgb]{ .961, .985, .970 }
& \textsc{DDCoT} \neurips{NeurIPS 2023} & 43.9 & 45.3 & 62.8 & 65.3 & 1752.4 & 1826.6 \\
\rowcolor[rgb]{ .961, .985, .970 }
& \textsc{SCAFFOLD} \acl{ACL 2025} & 49.9 & 53.6 & 74.4 & 75.0 & 1668.2 & 1822.2 \\
\rowcolor[rgb]{ .961, .985, .970 }
& \textsc{CCoT} \InputName{CVPR 2024} & 48.7 & 53.0 & 72.7 & 74.8 & 1866.3 & 1941.9 \\
\rowcolor[rgb]{ .961, .985, .970 }
& \textsc{ICoT} \InputName{CVPR 2025} & 38.0 & 44.8 & 54.2 & 67.0 & 1587.3 & 1709.3 \\
\rowcolor[rgb]{ .961, .985, .970 }
\multirow{-7.5}{*}{\shortstack{\textbf{\textit{Qwen2-VL-7B}} \\ \cite{wang2024qwen2}}} & \InitialReasoning{\textbf{\textsc{DaP-ICoT}}} & \textbf{57.2} & \textbf{58.7} & \textbf{75.9} & \textbf{78.5} & \textbf{2012.2} & \textbf{2076.0} \\

\bottomrule
\end{tabular}
\end{adjustbox}
\caption{The main experimental results. \textbf{Bold} indicates the best performance. For the M$^3$CoT and ScienceQA, Acc. is used as the evaluation metric, while for the MME, the sum of the Perception and Cognition scores is used as the evaluation metric.}
\vspace{-0.8\baselineskip}
\label{main results}
\end{table*}

\section{Experiments and Analysis}
\subsection{Experiments Setting}
Following \citet{gao2025interleaved}, in addition to the direct query approach, we also adopt the following methods as baselines:

\begin{itemize}
    \item \texttt{MMCoT}~\cite{mcot} separates rationale generation and answer inference by incorporating both text and image modalities to improve reasoning performance.
    
    \item \texttt{DDCoT}~\cite{ddcot} divides reasoning and recognition by combining LLM reasoning with visual recognition through negative-space prompting, enabling effective and explainable multimodal CoT reasoning.
    
    \item \texttt{SCAFFOLD}~\cite{scaffold} prompts overlays a dot matrix on images as visual anchors and introduces coordinate-based textual references to effectively enhance vision-language coordination in MLLMs.
    
    \item \texttt{CCoT}~\cite{ccot} first generates scene graphs with LMMs and then uses them in prompts to enhance compositional reasoning without annotations.
    
    \item \texttt{ICoT}~\cite{gao2025interleaved} generates paired visual and textual reasoning steps by inserting image regions via Attention-driven Selection to enhance reasoning.
\end{itemize}
\noindent
All methods are reproduced once using their official open-source implementations and evaluated under both 0-shot and 1-shot settings. To evaluate the effectiveness of \textsc{DaP-ICoT}, we conduct experiments on five MLLMs, including Chameleon-7B~\cite{team2024chameleon}, LLaVA-V1.5-(7B, 13B)~\cite{liu2023visual}, and Qwen2-VL-(2B, 7B)~\cite{wang2024qwen2}. We use the default top-p and temperature settings provided by each MLLM. In \textsc{DaP-ICoT}, the confidence threshold $\tau$ is tuned on M$^3$CoT validation set by searching within [0, 1] and selecting the value that yields the best performance.

\subsection{Main Results}
The experimental results are summarized in Table~\ref{main results}. Based on these results, we can observe the following:

\begin{itemize}
    \item [(1)] \textbf{\textsc{DaP-ICoT} achieves consistently superior performance.}
\textsc{DaP-ICoT} consistently achieves the highest reasoning accuracy across all settings. In particular, it significantly outperforms all baseline methods on the M$^3$CoT benchmarks under both 0-shot and 1-shot settings. For example, on M$^3$CoT task with the Chameleon-7B, \textsc{DaP-ICoT} achieves a remarkable 0-shot accuracy of 41.0\%, which is substantially higher than the second-best method, Scaffold, with a score of 31.0\%.

\item [(2)] \textbf{\textsc{DaP-ICoT} demonstrates strong versatility across diverse tasks.}
In addition to its promising performance on M$^3$CoT, \textsc{DaP-ICoT} consistently outperforms all baselines across other challenging reasoning tasks and comprehensive multimodal benchmarks. Specifically, it achieves the highest average scores on ScienceQA and MME in all settings. This demonstrates its strong generalization ability and adaptability to complex reasoning and holistic multimodal understanding tasks.

\item [(3)] \textbf{\textsc{DaP-ICoT} is generalizable across MLLMs of different architectures and scales.}
\textsc{DaP-ICoT} consistently delivers superior performance across various MLLMs with diverse architectures and scales. From small MLLMs such as \textit{Qwen2-VL-2B} to large MLLMs like \textit{Qwen2-VL-7B} and \textit{LLaVA-V1.5-13B}, \textsc{DaP-ICoT} maintains its leading performance. These results clearly suggest that \textsc{DaP-ICoT} is not only effective for specific models but also adaptable to a wide range of pretraining settings and reasoning capacities, demonstrating scalability and model-agnostic robustness.
\end{itemize}

\subsection{Analysis}
This section provides a more in-depth analysis of \textsc{DaP-ICoT} to demonstrate its effectiveness and efficiency.

\subsubsection{1. Both the DVTI and PVTG modules are vital for addressing key ICoT challenges.} 

\begin{figure}[!t]
\centering
\includegraphics[width=\columnwidth]{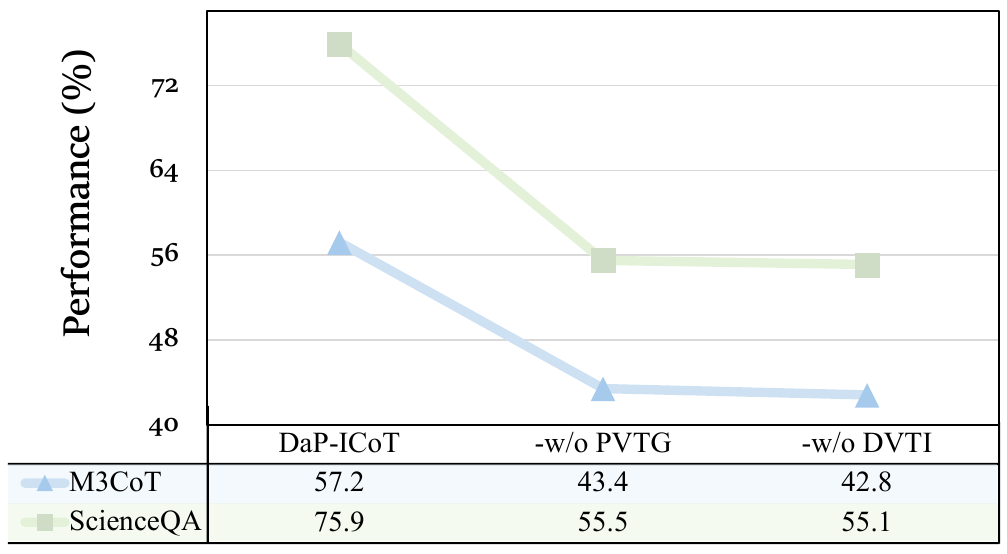}
\caption{Ablation Study on \textit{Qwen2-VL-7B}: ``w/o PVTG'' indicates removal of \textit{Precise Visual Thought Guidance} for Visual Cues, and ``w/o DVTI'' indicates removal of \textit{Dynamic Visual Thought Integration} for Adaptive Reasoning}
\label{ablation}
\end{figure}

We perform thorough ablation studies on \textit{Qwen2-VL-7B} to systematically assess the impact of \textit{Dynamic Visual Thought Integration} (DVTI) and \textit{Precise Visual Thought Guidance} (PVTG). The results, as shown in Figure~\ref{ablation}, clearly confirm their effectiveness.

\begin{itemize}
    \item Removing the \textit{Dynamic Visual Thought Integration} (DVTI) module results in a substantial performance degradation—specifically, a 14.4\% drop on the M$^3$CoT benchmark and a 20.8\% drop on the ScienceQA benchmark. These results underscore the critical role of DVTI in dynamically fusing multimodal information. 

    \item Removing the \textit{Precise Visual Thought Guidance} (PVTG) module leads to a significant performance decline, with a 13.8\% reduction on the M$^3$CoT benchmark and 20.4\% on the ScienceQA benchmark. This highlights the importance of PVTG in structuring the visual input space by leveraging fine-grained object-level segmentation. 
\end{itemize}

These findings demonstrate that both DVTI and PVTG are essential for efficient and precise ICoT reasoning.

\begin{figure}[!t]
\centering
\includegraphics[width=\columnwidth]{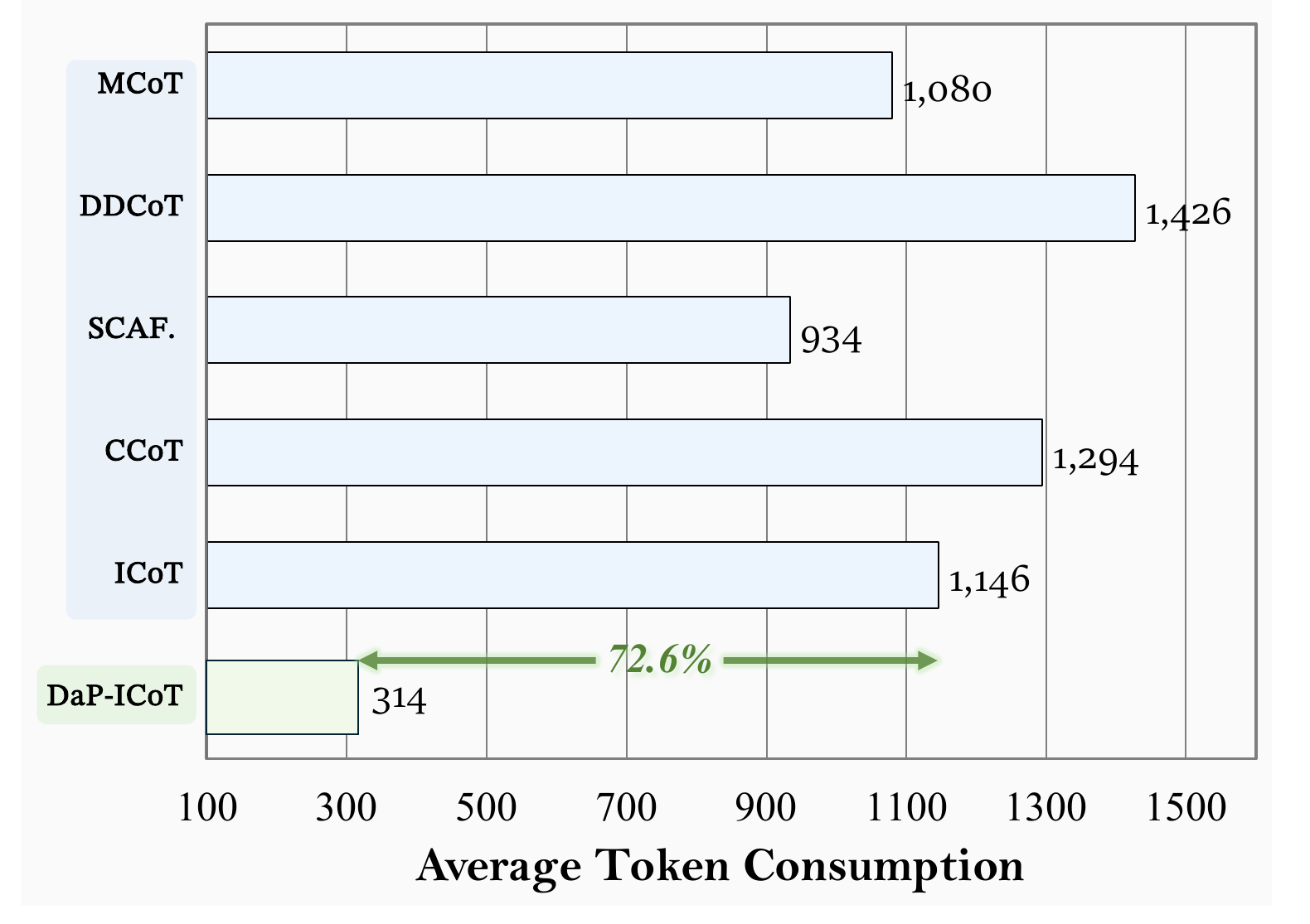}
\caption{A comparison of the total token consumption between \textsc{DaP-ICoT} and baseline methods on the M$^3$CoT benchmark using the \textit{Qwen2-VL-7B}. \textsc{DaP-ICoT} achieves a 72.6\% reduction in token consumption compared to ICoT.}
\label{token}
\end{figure}

\subsubsection{2. \textsc{DaP-ICoT} significantly reduces the token consumption of MLLMs.}
To further verify the lightweight design of \textsc{DaP-ICoT}, we compare its total token consumption with several baseline methods on the M$^3$CoT using the \textit{Qwen2-VL-7B}. This experiment aims to assess whether \textsc{DaP-ICoT} can effectively reduce token usage while maintaining strong reasoning performance. As shown in Figure~\ref{token}, \textsc{DaP-ICoT} achieves a significant reduction in token consumption, using an average of 314 tokens, which is a 72.6\% decrease compared to \texttt{ICoT} that requires 1,146 tokens. Furthermore, most baselines consume considerably more tokens, with \texttt{CCoT} requiring 1,294 tokens and \texttt{DDCoT} reaching the highest at 1,426 tokens. Even relatively more efficient methods, such as \texttt{MMCoT} and \texttt{SCAFFOLD}, still require 1,080 and 934 tokens, respectively. These results indicate that the \textit{Dynamic Visual Thought Integration} module in \textsc{DaP-ICoT} effectively reduces the overhead associated with image embedding, thereby achieving significantly lower token consumption compared to existing baseline reasoning methods. 

\begin{figure}[!t]
\centering
\includegraphics[width=\columnwidth]{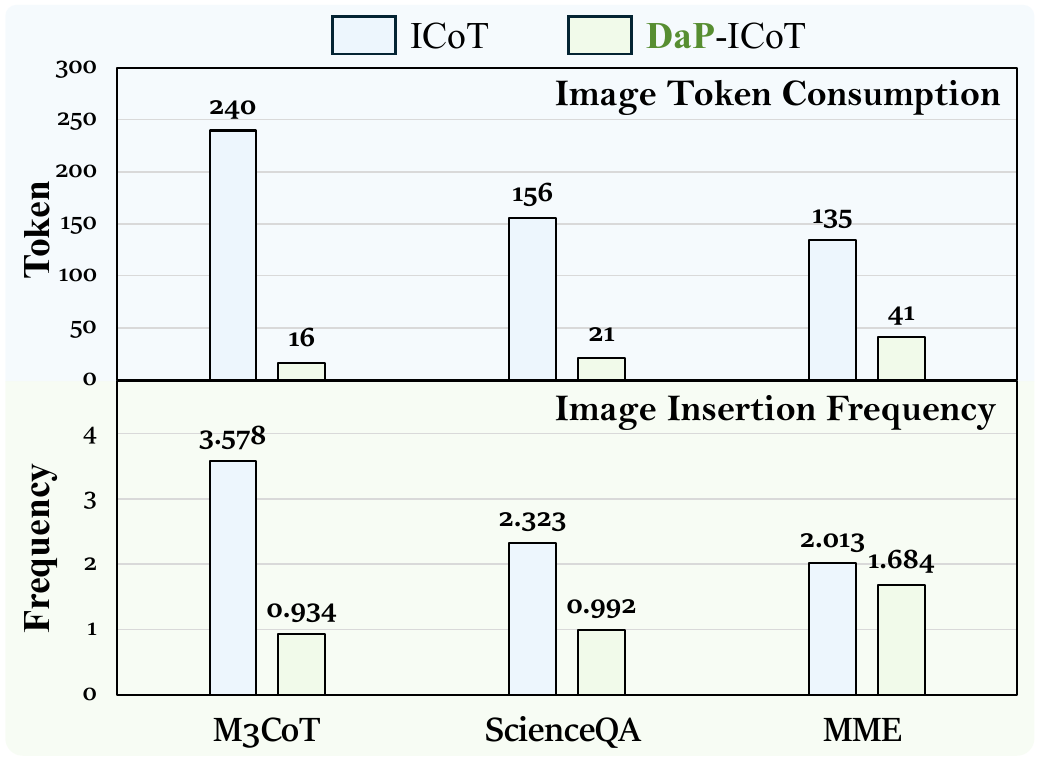}
\caption{A comparison of image insertion frequency and the number of inserted image tokens between \textsc{DaP-ICoT} and ICoT on the \textit{Qwen2-VL-7B} model.}
\label{frequency}
\end{figure}

\begin{figure}[!t]
\centering
\includegraphics[width=\columnwidth]{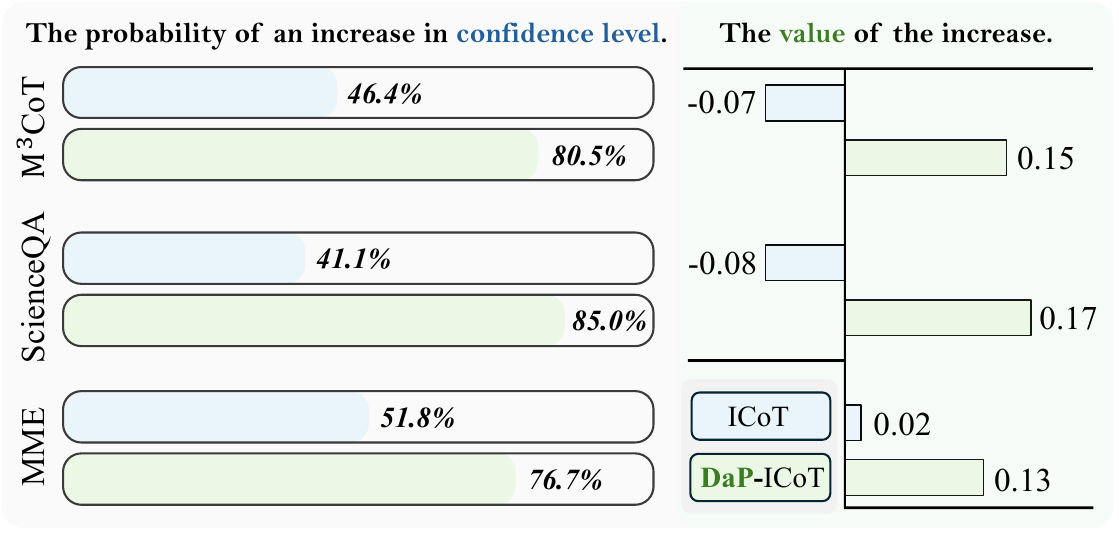}
\caption{The proportion of samples where the confidence level increases after image insertion, as well as the average confidence value improvement, for \textsc{DaP-ICoT} and ICoT. The \Video{Blue Color} denotes the baseline method ICoT, while the \InitialReasoning{Green Color} represents our \textsc{DaP-ICoT}.}
\label{proportion}
\end{figure}

\subsubsection{3. \textsc{DaP-ICoT} reduces resource consumption from image insertions.}
To clarify the efficient feature of \textsc{DaP-ICoT}, we conduct a comparative analysis with ICoT on image usage during reasoning, based on two key metrics: (1) The average number of image insertions per sample and (2) The average number of image tokens consumed after insertion. As shown in Figure~\ref{frequency}, \textsc{DaP-ICoT} demonstrates a significant reduction in both image insertion frequency and token consumption. On average, \textsc{DaP-ICoT} inserts only 1.2 images per sample, whereas ICoT inserts an average of 2.6 images. Moreover, in terms of token usage, \textsc{DaP-ICoT} consumes merely 26 image tokens on average, which is substantially lower than ICoT. These results demonstrate the efficiency of \textsc{DaP-ICoT}, which achieves superior performance with minimal visual input and token consumption. This is due to its modules: \textit{Dynamic Visual Thought Integration}, which adaptively reduces unnecessary image insertions, and \textit{Precise Visual Thought Guidance}, which effectively minimizes token usage by selecting compact object-level visual inputs.

\begin{figure}[!t]
\centering
\includegraphics[width=\columnwidth]{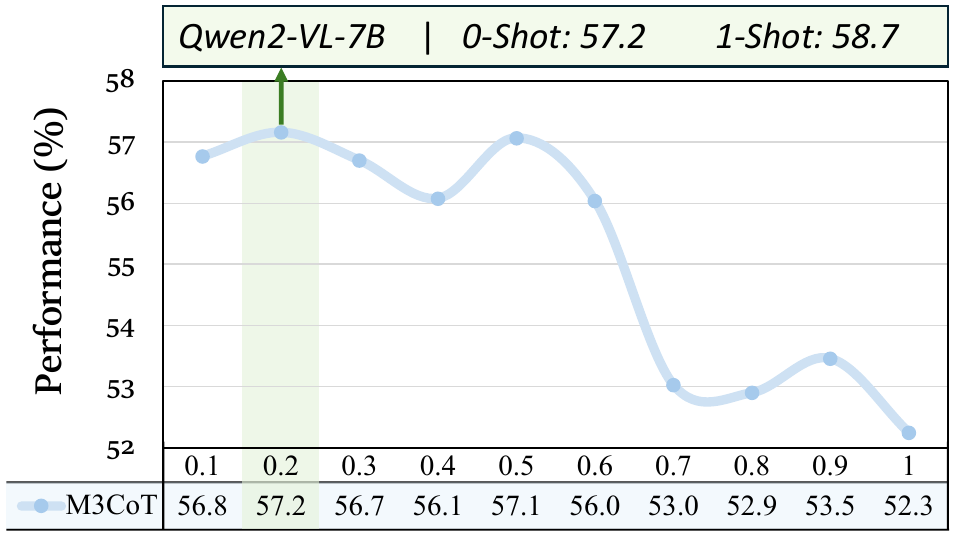}
\caption{Performance comparison across the Confidence Threshold $\tau$ range of $(0, 1]$: Optimal model performance achieved at $\tau = 0.2$, with 57.2\% Accuracy in the 0-shot setting and 58.7\% in the 1-shot setting, effectively balancing visual integration and textual reasoning.}
\label{confidence}
\end{figure}

\subsubsection{4. \textsc{DaP-ICoT} effectively enhances confidence during the reasoning process.}
To further investigate why \textsc{DaP-ICoT} achieves superior performance, we conduct an in-depth analysis of the model's reasoning confidence variations after image insertion. Specifically, we compare \textsc{DaP-ICoT} and ICoT in terms of two key metrics: (1)~The proportion of samples showing increased confidence after applying Visual Thought, and (2) The average magnitude of the confidence improvement. As shown in Figure~\ref{proportion}, \textsc{DaP-ICoT} consistently demonstrates a significantly higher probability of confidence improvement across all three benchmarks. On average, \textsc{DaP-ICoT} leads to an increase in confidence for 80.7\% of the samples, while ICoT only improves confidence in 46.4\% of cases. Furthermore, regarding the extent of confidence enhancement, \textsc{DaP-ICoT} consistently outperforms ICoT across all three benchmarks. These results clearly indicate that \textsc{DaP-ICoT} is more effective in leveraging visual information to enhance confidence during reasoning, thereby contributing to its superior performance.

\begin{figure*}[t]
\centering
\includegraphics[width=0.9\textwidth]{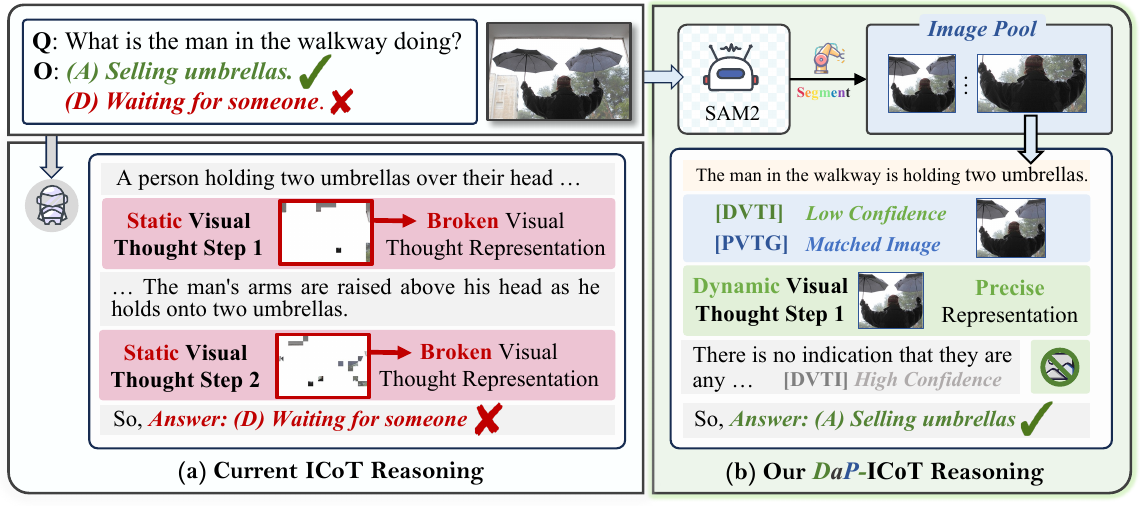} 
\caption{The case study. Figure (a) illustrates the reasoning process of the current ICoT, where after each reasoning step, it is \textit{\textbf{necessary}} to insert image tokens that are most similar to the previous text. However, these tokens are \textit{\textbf{broken}} image tokens, which ultimately leads to the incorrect answer \textit{(D)}. Figure (b) illustrates our \textsc{DaP-ICoT}. In contrast, \textsc{DaP-ICoT} \textit{\textbf{only}} inserts images when the text confidence is low. Furthermore, the inserted images are \textit{\textbf{precise and complete}}, having been segmented using the SAM2 model. After efficient interleaved visual-textual reasoning, the final correct answer \textit{(A)} is obtained.}
\label{case}
\end{figure*}

\subsubsection{5. The search strategy for the confidence threshold $\tau$.}
To gain a clearer understanding of the threshold selection process for the confidence threshold $\tau$ in the \textit{Dynamic Visual Thought Integration} (DVTI) module, we conduct a systematic threshold search experiment using \textit{Qwen2-VL-7B} on the M$^3$CoT benchmark. Specifically, we vary the confidence threshold $\tau$ within the range of $(0, 1]$ with an interval of 0.1 and evaluate the model’s performance under each setting. The experimental results are presented in Figure~\ref{confidence}, illustrating the relationship between the confidence threshold and overall performance. Reveal the impact of varying the confidence threshold $\tau$ on model performance. As $\tau$ increases, the performance exhibits a clear trend. The optimal performance of 57.2\% is achieved at a threshold of 0.2. This trend suggests that a moderate threshold effectively balances the model's reliance on visual information and textual reasoning, enabling optimal reasoning performance. In contrast, overly conservative or overly aggressive visual interactions negatively affect the model’s ability to reason effectively.

\subsubsection{6. Qualitative Analysis.}
To better understand the performance of \textsc{DaP-ICoT}, we present a real-world example. As shown in Figure~\ref{case} (a), ICoT inserts incomplete image tokens during reasoning, leading to an incorrect result (\textit{D}). This example highlights the limitations of the current ICoT approach and its impact on reasoning accuracy. In contrast, Figure~\ref{case} (b) demonstrates \textsc{DaP-ICoT}, which inserts only complete, context-relevant images when text confidence is low. Compared to ICoT, \textsc{DaP-ICoT} inserts \textit{fewer} images and employs a more selective approach. After ICoT reasoning, the system correctly arrives at answer (\textit{A}), which is verified through the dynamic visual thought integration mechanism. This example illustrates the efficiency of the selective visual input mechanism in \textsc{DaP-ICoT} reasoning.

\section{Related Work}
In recent years, Multimodal Large Language Models (MLLMs) have witnessed rapid advancements~\cite{liang2024survey,qiu2025intentvcnet,qiang2025ver}, and the emergence of Multimodal Chain-of-Thought (MCoT) reasoning has further enhanced their performance~\cite{wang2025multimodal}.
Specifically, \citet{mcot} proposed Multimodal-CoT, a two-stage reasoning framework integrating text and image modalities.
\citet{chen-etal-2024-m3cot} introduce M$^3$CoT, a benchmark for multi-modal, multi-step, and multi-domain chain-of-thought reasoning, addressing key limitations of existing MCoT benchamrk. However, MCoT methods largely follow the conventional paradigm of taking cross-modal inputs while generating reasoning outputs only in the text modality. This limits the effective use of modality complementarity and diminishes reasoning performance~\cite{wang2025vision,lin2025plan,zhang2025cchall}.

To address this limitation, researchers have explored Interleaved-Modal Chain-of-Thought (ICoT)~\cite{gao2025interleaved,wu2025vic}, which enhances the reasoning abilities of MLLMs through cross-modal Integrations \cite{hu2024visual,cheng2025visual}. For example, \citet{zhou2024image} propose Image-of-Thought prompting to guide MLLMs in step-by-step visual extraction. \citet{hu2024visual} introduce a sketching framework, enabling models to perform human-like drawing to reasoning. In addition, \citet{cheng2025comt} propose the CoMT for evaluating multimodal reasoning with visual and textual operations. \citet{gao2025interleaved} introduce ICoT reasoning with an Attention-driven Selection for generating interleaved visual-textual reasoning.

Compared to previous ICoT reasoning approaches, \textsc{DaP-ICoT} introduces both \textit{Dynamic Visual Thought Integration} and \textit{Precise Visual Thought Guidance}, enabling not only more efficient reasoning but also adaptive and context-aware visual clues for ICoT Reasoning.

\section{Conclusion}
In this work, we propose Interleaved-modal Chain-of-Thought reasoning with \textbf{D}ynamic \textbf{a}nd \textbf{P}recise Visual Thoughts (\textsc{DaP-ICoT}), achieving efficient reasoning. Specifically, \textsc{DaP-ICoT} adaptively integrates informative and context-relevant visual information and provides semantically coherent visual inputs. Extensive evaluations on multiple benchmarks and advanced MLLMs demonstrate that \textsc{DaP-ICoT} achieves superior performance. In addition, \textsc{DaP-ICoT} is capable of effectively reducing token consumption and the frequency of visual insertions, highlighting its strong potential in efficient multimodal reasoning.

\section*{Acknowledgments}
This work was supported by the National Natural Science Foundation of China (NSFC) via grants 92570120 and 62306342. This work was supported by the Scientific Research Fund of Hunan Provincial Education Department (24B0001). This work was sponsored by the Excellent Young Scientists Fund in Hunan Province (2024JJ4070), the Science and Technology Innovation Program of Hunan Province under Grant 2024RC3024, and CCF-Zhipu Large Model Innovation Fund (NO.CCF-Zhipu202406). This study was also funded by the Open Project of the Text Computing and Cognitive Intelligence Ministry of Education Engineering Research Center (No. TCCI250101). Libo Qin is the corresponding author.

\bibliography{aaai2026}

\end{document}